\documentclass[runningheads]{llncs}
\usepackage[T1]{fontenc}
\usepackage{graphicx}
\usepackage{amsmath}  
\usepackage{amssymb}  
\usepackage{bm}       

\usepackage{cite}

\usepackage{booktabs}
\begin{document}
\title{A Review of Hybrid and Ensemble in Deep Learning for Natural Language Processing}
%
%
\author{Jianguo Jia\inst{1} \and
Wen Liang \inst{2}\orcidID{0009-0006-5646-2214} \and
Youzhi Liang\inst{3}}
\authorrunning{J. Jia et al.}
\titlerunning{A Review of Hybrid and Ensemble in NLP}

%
\institute{Hong Kong Polytechnic University, Hong Kong, China \email{jianguo1.jia@connect.polyu.hk}\and
Google Research, Mountain View CA 94043, USA \email{liangwen@google.com} \and
Stanford University, Stanford CA 94305, USA \email{youzhil@stanford.edu}}

%
\maketitle              

\begin{abstract}
This review presents a comprehensive exploration of ensemble deep learning models within Natural Language Processing (NLP), shedding light on their transformative potential across diverse tasks such as Sentiment Analysis, Named Entity Recognition, Machine Translation, Question Answering, Text Classification, Generation, Speech Recognition, Summarization, and Language Modeling. The paper systematically introduces each task, delineates key architectures from Recurrent Neural Networks (RNNs) to Transformer-based models like BERT and large language models, and evaluates their performance, challenges, and computational demands. The adaptability of ensemble techniques is emphasized, highlighting their capacity to enhance various NLP applications. Challenges in implementation, including computational overhead, overfitting, and model interpretation complexities, are addressed, alongside the trade-off between interpretability and performance. Serving as a concise yet invaluable guide, this review synthesizes insights into tasks, architectures, and challenges, offering a holistic perspective for researchers and practitioners aiming to advance language-driven applications through ensemble deep learning in NLP.


\end{abstract}

\section{Introduction}
Natural Language Processing (NLP) \cite{jones1994natural}\cite{liddy2001natural}, an interdisciplinary field combining elements of computational linguistics, artificial intelligence, and computer science, aims to enable machines to understand, interpret, and generate human language \cite{jurafsky2014speech}. Originating in the 1950s with endeavors such as the Georgetown experiment \cite{hutchins1995machine}\cite{nadkarni2011natural}\cite{turing1950mind}, NLP has undergone significant transformations. Initially relying on rule-based systems, the field gradually shifted towards data-driven methodologies with the integration of statistical models in the late 20th century \cite{manning2008introduction}. This shift paved the way for contemporary advancements brought about by deep learning techniques.

Deep learning, a subset of machine learning, involves training neural networks on large datasets to perform tasks without task-specific programming \cite{deng2014deep}\cite{zhang2021understanding}\cite{goodfellow2016deep}. It has emerged as a pivotal force in NLP, revolutionizing the field with models such as recurrent neural networks (RNNs), convolutional neural networks (CNNs), and transformers \cite{vaswani2017attention}. Techniques like Word2Vec \cite{mikolov2013efficient} have been instrumental in creating word embeddings, capturing semantic relationships in continuous vector spaces and serving as a basis for more advanced models.

The application of deep learning in NLP has yielded several breakthroughs. It has enabled the automatic learning of complex patterns in large datasets, enhancing the accuracy and performance across various tasks. The capacity for transfer learning allows models pre-trained on substantial datasets to be fine-tuned for niche tasks with limited data \cite{howard2018universal}. Unlike conventional methods requiring manual feature engineering, deep learning models autonomously discern features and hierarchies within data, resulting in more robust models \cite{young2018recent}. The advent of architectures such as BERT \cite{devlin2019bert} has underscored the supremacy of deep learning in NLP by establishing new benchmarks.

Large Language Models (LLMs) have shown exceptional capabilities in addressing a wide range of tasks within various NLP domains. These models, such as GPT-3 and Gemini \cite{brown2020languagemodelsfewshotlearners}, have revolutionized fields like machine translation, text summarization, and question answering, demonstrating their versatility and effectiveness \cite{brown2020languagemodelsfewshotlearners}. Recent studies have further explored the potential of LLMs in more specialized applications, such as legal document analysis and medical diagnostics, highlighting their adaptability and potential for real-world impact \cite{colombo2024saullm7bpioneeringlargelanguage,cui2024chatlawmultiagentcollaborativelegal}. However, despite their impressive performance, challenges such as ethical considerations, computational costs, and the need for large datasets remain significant areas of ongoing research and development \cite{bender2021dangers}.

In this context, the exploration of hybrid and ensemble deep learning approaches becomes crucial. These strategies aim to further augment the capabilities of individual models by combining their strengths and mitigating their limitations. This paper provides a comprehensive review of such approaches, delineating the progress and the challenges faced in the ongoing journey of NLP.

Ensemble methods \cite{dietterich2000ensemble}\cite{opitz1999popular} involve combining multiple models to improve the overall performance of a machine-learning task. The underlying principle is that leveraging the strengths and mitigating the weaknesses of multiple models can achieve better accuracy and robustness than any single model \cite{zhou2012ensemble}. Ensemble learning encompasses a range of approaches, including Bagging, Boosting, and Stacking \cite{mi2019bagging}\cite{kumar2008from}, which have gained significant popularity in the field. The technique known as bagging, short for Bootstrap Aggregating, entails the training of numerous iterations of a single model on distinct subsets of the training data, followed by the averaging of their respective predictions \cite{breiman1996bagging}. Boosting, on the other hand, trains models sequentially, where each new model attempts to correct the errors made by its predecessors \cite{freund1997decision}. Stacking involves training multiple models and using another model, called a meta-learner, to combine their predictions \cite{wolpert1992stacked}.

Ensemble methods \cite{dietterich2000ensemble}\cite{opitz1999popular} in machine learning aim to enhance predictive performance by strategically combining several models. This synergistic approach often leads to improved accuracy and robustness compared to individual models, as it capitalizes on the strengths while offsetting the weaknesses of each model \cite{zhou2012ensemble}. Traditional ensemble techniques, such as Bagging, Boosting, and Stacking \cite{mi2019bagging}\cite{kumar2008from}, have been widely adopted. Bagging, or Bootstrap Aggregating, involves training multiple instances of a single model on different subsets of the data and then averaging their predictions to minimize variance \cite{breiman1996bagging}. In contrast, Boosting focuses on sequential model training, with successive models aiming to rectify the errors of their predecessors \cite{freund1997decision}. Stacking combines the predictions from multiple trained models using a secondary model, known as a meta-learner \cite{wolpert1992stacked}.

In the realm of Natural Language Processing (NLP), the integration of deep learning with ensemble methods has led to the emergence of hybrid and ensemble deep learning approaches. These approaches combine different neural network architectures or integrate deep learning models with traditional machine learning algorithms. For instance, an ensemble could include a combination of convolutional neural networks (CNNs), recurrent neural networks (RNNs), and transformers, each contributing to capturing different linguistic features and patterns.

The incorporation of hybrid and ensemble techniques in NLP aims to further elevate the performance metrics across tasks by leveraging the complementary strengths of diverse models. Ensemble methods can enhance the generalization capabilities of deep learning models, ensuring more consistent performance across various datasets and domains. Hybrid models, which may fuse deep learning techniques with traditional machine learning approaches, strive to amalgamate the representational power of neural networks with the interpretability and simplicity of classical algorithms.

This paper delves into the intricacies of hybrid and ensemble deep learning approaches in NLP, exploring their conceptual foundations, applications, and the challenges and opportunities they present in the quest for enhanced language understanding and processing.

\section{Base Models}
This section delves into the base models employed in Natural Language Processing (NLP). These foundational models act as essential components that will be incorporated into the hybrid or ensemble models discussed in the subsequent section. We start with classic machine learning models, then dive deeper into deep learning models and large language models.

\subsection{Recurrent Neural Networks (RNN) in Sentiment Analysis}
Recurrent Neural Networks (RNNs) have significantly influenced the field of sentiment analysis within natural language processing (NLP), as evidenced by various studies \cite{kurniasari2020sentiment}\cite{arras2017explaining}\cite{patel2019sentiment}\cite{tembhurne2021sentiment}\cite{alhumoud2022arabic}\cite{mao2022review}. Characterized by their ability to process sequential data, RNNs excel at capturing temporal information and context, which are crucial for understanding sentiment in text \cite{elman1990finding}\cite{graves2012supervised}. Unlike traditional neural networks, RNNs process inputs sequentially, maintaining a 'memory' of previous inputs within their internal state or hidden layer, allowing for effective sequential information processing \cite{lipton2015critical}. The hidden state of an RNN is determined by both the current input and the prior hidden state, thereby enabling the network to accumulate information over successive time steps \cite{ming2017understanding}. For training, RNNs employ Backpropagation Through Time (BPTT), a technique that involves unrolling the network over time before applying the standard backpropagation algorithm \cite{werbos1990backpropagation}.

RNNs' ability to understand context within sentences and paragraphs is pivotal for accurate sentiment analysis. Nevertheless, they sometimes encounter difficulties with extended sequences due to issues such as the vanishing gradient problem. Long Short-Term Memory (LSTM) networks and Gated Recurrent Units (GRUs), which are advanced variations of RNNs, often outperform basic RNNs in sentiment analysis tasks by providing improved accuracy and better management of dependencies \cite{zhang2018character}\cite{baktha2017investigation}\cite{sachin2020sentiment}\cite{islam2020review}. Moreover, ensemble or hybrid approaches, which combine different models or algorithms, have been explored to further boost the performance of sentiment analysis.

Among the strengths of RNNs is their enhanced contextual understanding. They are proficient at capturing and processing sequential information within texts, interpreting not just individual elements in isolation but also their interrelations within a broader narrative \cite{baktha2017investigation}\cite{maheswaranathan2020how}\cite{ito2020contextual}. This trait makes RNNs effective for tasks such as sentiment analysis and language translation. Another advantage is their capability to handle variable-length inputs seamlessly. RNNs are inherently designed to accommodate input sequences of varying lengths, which is essential in NLP where sentence lengths can differ considerably \cite{baziotis2017datastories}.

However, RNNs also present certain challenges. One of the primary concerns is the high computational demand during training, particularly for long sequences. The step-by-step processing nature of RNNs necessitates frequent weight updates, becoming increasingly resource-intensive with longer sequences and more complex models. Additionally, the use of BPTT for training increases the computational load significantly, especially for long sequences. Another fundamental problem faced by RNNs is the vanishing gradient issue, where gradients propagated back in time tend to diminish in long sequences, hindering the network's ability to learn long-range dependencies \cite{birjali2021comprehensive}.

\subsection{Convolutional Neural Networks (CNN) in Sentiment Analysis}

Convolutional Neural Networks (CNNs) have gained widespread acclaim for their capabilities in image processing and computer vision, and their contributions have extended significantly to the field of Natural Language Processing (NLP), particularly in the specialized arena of sentiment analysis\cite{gandhi2021sentiment}\cite{rani2019deep}\cite{yu2016visual}\cite{kim2019sentiment}. These networks are tailored for recognizing local patterns and features, rendering them particularly apt for identifying phrases and patterns in textual data that are indicative of sentiment\cite{kim2014convolutional}. The distinctive architecture of CNNs is marked by the integration of convolution layers, pooling layers, and fully connected layers, each contributing to the nuanced task of sentiment detection and classification.

In sentiment analysis, CNNs employ convolution layers to apply a series of filters to input text, thereby extracting features with semantic significance. These filters traverse word embeddings, diligently capturing the semantic nuances from contiguous word groupings. Subsequent pooling layers, predominantly max pooling, distill this information, reducing dimensionality and emphasizing the most prominent features. The extracted features are subsequently flattened and navigated through fully connected layers, culminating in the final sentiment classification\cite{lecun2015deep}.

CNNs have proven themselves to be exemplary in hierarchical feature learning within sentiment analysis. From identifying simple word embeddings to extracting complex semantic structures across extended text sequences, their capability is prominently displayed in tasks such as discerning sentiment nuances in phrases like "not good". CNNs adeptly sift sentiment-laden features from broader contexts, ensuring that subtle expressions that might be neglected by alternative models are captured. The versatility of CNNs is further underscored by their ability to adapt to texts of varying lengths through the use of kernels of different sizes. This enables the comprehension of both succinct phrases and extensive contextual sequences. Pooling layers enhance this adaptability, ensuring that poignant sentiment indicators are consistently focused upon, leading to standardized outputs and consistent sentiment evaluations.

The efficacy of CNNs in sentiment analysis is highlighted by their efficiency and reduced complexity. They often outperform certain recurrent models in classification accuracy and training efficiency, as demonstrated by Zhang in 2015\cite{zhang2015character}. Their versatility spans a range of sentiment analysis applications, from binary to fine-grained sentiment classifications, thus establishing CNNs as a comprehensive and effective tool in sentiment analysis.

CNNs are renowned for their efficiency, particularly when juxtaposed against Recurrent Neural Networks (RNNs). The architecture of CNNs facilitates parallel processing, allowing concurrent convolution operations across text segments. This feature ensures reduced training times for large datasets and presents a distinct advantage over sequentially operating RNNs. Modern hardware optimizations for matrix operations, as seen in Graphics Processing Units (GPUs), further amplify the efficiency of CNNs during training and inference.

In terms of feature extraction, CNNs exhibit exemplary capabilities. Their architecture excels in identifying position-invariant patterns within text, ensuring consistent capture of sentiment-rich phrases. The hierarchical structure of CNNs facilitates a tiered approach to feature detection, negating the need for labor-intensive manual feature engineering prevalent in traditional sentiment analysis.

However, CNNs are not without their limitations. A notable challenge is their sensitivity to context, or rather, the lack thereof. Unlike RNNs or Transformers, CNNs may occasionally falter in comprehending extended contextual relationships within textual data. Another limitation arises from the employment of fixed-size kernels, determining the contextual window from which the network learns. The selection of kernel size is critical and demands careful consideration, as it may impact the network's ability to understand broader contexts and potentially restrict its effectiveness in complex sentiment analysis tasks.

\subsection{Long Short-Term Memory Networks (LSTM)}
Long Short-Term Memory (LSTM) networks, a specialized variant of Recurrent Neural Networks (RNNs), have dramatically influenced natural language processing (NLP) and sentiment analysis due to their capacity to encapsulate long-term dependencies in sequences\cite{miedema2018sentiment}\cite{shobana2021efficient}\cite{al2019using}\cite{muhammad2021sentiment}\cite{yadav2023habitat}. These networks are built around memory cells, which, akin to computer memory, are adept at reading, writing, and preserving crucial information across arbitrary time intervals, ensuring the retention of contextually pertinent data\cite{staudemeyer2019understanding}. The flow of information within LSTMs is meticulously regulated by a series of gates: the forget gate, which selectively retains or discards information from the cell state; the input gate, which decides the values to be updated in the cell state and computes new information; and the output gate, which determines the next hidden state by deciding what information from the current cell state is outputted\cite{olah2015understanding}.

LSTMs were developed as an antidote to the vanishing gradient problem that plagued traditional RNNs, which tended to forget distant information in sequences. By facilitating the effective flow of gradients over extended sequences, LSTMs are capable of learning and retaining information from early inputs while processing subsequent ones, a monumental leap over conventional RNNs\cite{fei2018bidirectional}\cite{hassan2017deep}\cite{hochreiter1997long}. In sentiment analysis, the LSTM's ability to preserve and comprehend the broader context is paramount. For instance, they can seamlessly bridge gaps in context, providing a comprehensive understanding of sentiments even in complex sentences where crucial contextual clues may be interspersed with less relevant information. Additionally, LSTMs are proficient at discerning the subtle shifts in sentiment based on the structure and arrangement of words within a sentence.

LSTMs have established themselves as formidable tools in sentiment analysis by consistently achieving benchmark results on various datasets. Furthermore, their versatility extends to integration with other neural network architectures to create ensemble or hybrid models. For example, combining LSTMs with Convolutional Neural Networks (CNNs) creates a potent system wherein LSTMs analyze sequences and context while CNNs excel at local feature extraction. This synergistic approach has been proven to enhance performance in sentiment analysis tasks\cite{Yin2017}.

LSTMs' proficiency in capturing long-range dependencies and their adaptability to handle variable input sequences make them remarkably flexible across applications such as sentiment analysis, machine translation, time series prediction, and music generation. However, they are not without drawbacks. The computational intensity of training LSTMs, particularly on extensive datasets, can be taxing, necessitating powerful hardware for expedited processing. Moreover, LSTMs are susceptible to overfitting, often requiring regularization techniques to ensure that the model generalizes effectively to unseen data. Overall, LSTMs, despite their few limitations, continue to be a cornerstone in the realm of sentiment analysis, providing nuanced and contextually rich interpretations of textual data.

\subsection{Bidirectional Encoder Representations from Transformers (BERT)}
Bidirectional Encoder Representations from Transformers (BERT), introduced by Google in 2018, has significantly impacted the field of Natural Language Processing (NLP), particularly in applications such as sentiment analysis\cite{alaparthi2020bert}\cite{deepa2021bidirectional}\cite{devlin2018bert}. BERT is built upon the Transformer architecture, leveraging self-attention mechanisms to dynamically weigh different input tokens and discern relationships within a text. Unlike traditional NLP models that interpret text unidirectionally, BERT analyzes the context of a word bidirectionally, considering both preceding and following words. This approach ensures a holistic understanding of the context in which a word appears.

BERT's effectiveness is also attributed to its training methodology, where it undergoes pre-training on extensive datasets such as BooksCorpus and English Wikipedia. This comprehensive training equips BERT with a broad linguistic understanding, allowing it to capture general patterns which can then be fine-tuned for specific tasks, such as sentiment analysis. Its ability to understand text nuances is particularly beneficial in sentiment analysis, as it can detect idioms, colloquialisms, and sentiment-indicative phrases effectively.

In sentiment analysis, BERT demonstrates its adaptability through fine-tuning for task specificity. Initially trained on diverse datasets, BERT can be further tailored to specific sentiment analysis tasks, even with limited labeled data, ensuring high accuracy. BERT's bidirectional nature allows it to comprehend context and nuances, providing an advantage in sentiment analysis tasks involving complex textual data. It has consistently achieved state-of-the-art results across various sentiment analysis datasets, setting new performance benchmarks.

Furthermore, BERT is scalable, with larger models generally yielding better performance. However, this scalability often requires increased computational resources\cite{sun2019utilizing}. The architecture of BERT is designed to facilitate transfer learning, where the model is first trained on a large corpus of data to grasp linguistic patterns and then fine-tuned on smaller, task-specific datasets. This approach has led to significant performance improvements in NLP tasks. BERT's contextual representations also mark a departure from traditional word embedding methods, offering dynamic word representations based on context.

Despite its strengths, there are challenges associated with BERT. The model, particularly its larger variants, is resource-intensive, necessitating substantial computational power for both training and inference. This can be challenging for real-time applications or in scenarios with limited computational resources. Additionally, like many deep learning models, BERT operates as a "black box," where the internal processes leading to an output can be opaque. This lack of interpretability can be a concern in domains where understanding the rationale behind predictions is crucial.

Moreover, BERT's architecture allows for the exploration of ensemble or hybrid models, combining its strengths with other algorithms to further enhance its capabilities. For instance, integrating BERT with other machine learning models can potentially yield a system that not only captures contextual information efficiently but also addresses specific nuances or challenges presented by diverse datasets. Such hybrid models have the potential to further elevate the performance and applicability of BERT in sentiment analysis and other NLP tasks.

\subsection{Large Language Models}

Large Language Models (LLMs) have emerged as a transformative milestone in the field of Natural Language Processing (NLP), demonstrating remarkable abilities across a diverse array of tasks. LLMs like GPT-3 and Gemini have significantly advanced applications such as machine translation, text summarization, and question answering ~\cite{brown2020languagemodelsfewshotlearners}. In recent years, there have been significant efforts within the research community to create and share open-source LLMs. Projects like Hugging Face's Transformers library and EleutherAI's GPT-Neo have made powerful language models accessible to a broader audience, fostering an environment of shared knowledge and collective advancement ~\cite{kashyap2023gptneocommonsensereasoning}. LLAMA and Mistral have further advanced the research on open-source LLMs ~\cite{llama,mistral}. These open-source initiatives not only democratize the use of LLMs but also encourage continuous improvement and adaptation through community contributions. Certain models demonstrate superior overall performances on leaderboards like AlpacaEval and Chatbot Arena. However, is it reasonable to stick with one top-performing LLM for all user inputs? The answer may not be as straightforward as one might think. The LLM-Blender proposes an approach of multiple LLM ensemble ~\cite{jiang2023llmblenderensemblinglargelanguage}. It operates with two central modules, PairRanker and GenFuser, which are meticulously designed for executing the ranking and fusing stages correspondingly. During the ranking stage, it presents a specific input $x$ to $N$ different LLMs and compile their outputs as candidates ($y_1, \ldots, y_N$). A pairwise ranking module, PairRanker, is then employed to analyze and rank these candidates. Proceeding to the fusion stage, a sophisticated fusion model produces a final output considering the input and top $K$ candidates.

LLM-Blender introduces an ensemble method based on blending, which optimizes the generation process by combining the outputs from multiple models. However, LLM-Blender and similar methods need every LLM in the collections to generate a response, which can place a large amount of computational costs \cite{chen2023detectingerrorsestimatingaccuracy,jiang2023llmblenderensemblinglargelanguage}. Another  approach in the ensemble of LLMs is based on routing. This technique aims to dynamically route each input to the most suitable model, thereby enhancing both performance and efficiency. A notable example of LLM routing is FrugalGPT, which sequentially calls LLMs until a dedicated scoring model deems the generation acceptable ~\cite{chen2023frugalgptuselargelanguage}. This method optimizes resource usage by invoking simpler models first and escalating to more complex ones only when necessary. Prior works in this category necessitate training data that is sufficiently representative of each task and domain of interest to effectively train the ranking and scoring models.

Another significant contribution in this area is the Large Language Model Routing with Benchmark Datasets ~\cite{shnitzer2023largelanguagemodelrouting}. This approach involves learning LLM routers from benchmark datasets to ensure optimal performance when new tasks closely resemble the benchmark tasks. This strategy effectively reduces the out-of-distribution (OOD) gap, minimizing the need for extensive labeling efforts by practitioners. By leveraging benchmark datasets, the model routing system can generalize better to unseen tasks, providing a practical solution for diverse NLP applications without the additional labeling burden.

LLM serving costs are notably high. While the quality of the output is a significant aspect, another critical direction for optimization is reducing the costs. For instance, speculative decoding is a technique worth mentioning here ~\cite{leviathan2023fastinferencetransformersspeculative}. Although it does not aim to enhance generalizability or improve the overall performance like other ensemble methods, it is a good example of optimization efforts focused on speed using ensemble. Speculative decoding operates by generating multiple possible future tokens in parallel using a smaller or distilled model, then validating these tokens through the larger model. This approach can significantly reduce the inference time by preemptively exploring potential outputs, thus improving the efficiency of LLM serving without compromising the quality of the generated text. By integrating speculative decoding, we can achieve faster response times, making LLMs more practical for real-time applications and reducing the computational overhead associated with serving high-quality language models.

~\cite{chen2024cascadespeculativedraftingfaster}

\section{Hybrid and Ensemble Models for NLP}
This section delves into hybrid and ensemble models for NLP, exploring their applications across various tasks. The discussion is organized by NLP tasks, encompassing machine translation, question answering systems, named entity recognition, and language modeling.

\subsection{Machine Translation (MT)}
Machine Translation (MT) holds a pivotal position in computational linguistics, facilitating seamless communication across varied languages and cultures by autonomously translating text or speech from one language to another \cite{hutchins1995machine}\cite{niessen2000evaluation}\cite{kenny2018machine}\cite{sutskever2014sequence}. This domain has undergone remarkable evolutionary progress, driven by the continuous development and integration of statistical, neural, ensemble, and hybrid approaches.

Historically, the emergence of Statistical Machine Translation (SMT) in the late 1980s marked a significant transition. It continued to be a dominant approach through the 1990s and 2000s, fundamentally utilizing statistical models to deduce translation patterns from large bilingual corpora \cite{hearne2011statistical}\cite{lopez2008statistical}\cite{mahata2019mtil2017}\cite{doherty2012taking}\cite{brown1993mathematics}. The SMT framework comprises several components, including the Translation Model, which calculates the probability of translation between phrases in different languages, and the Language Model, which ensures the coherence of the generated text in the target language. The decoding step, an essential aspect of the process, involves optimizing to find the most probable translation. Alignment algorithms, such as the IBM Models, establish correspondence between source and target words during training \cite{brown1993mathematics}. Refinements in translation are achieved through phrase-based systems, reordering models, and model parameter tuning.

To augment the capabilities of SMT, ensemble methodologies were introduced. These techniques combine multiple models to produce more robust translations, proving particularly useful in addressing challenges associated with low-resource languages and domain adaptation. Through domain adaptation, SMT systems can be tailored to specific niches, such as medical or legal translations, ensuring precise translations of domain-specific terminology. Strategies like pivot translation, which involves using an intermediary language to bridge translation gaps, have proven beneficial for languages with scarce bilingual corpora.

With the advent of deep learning, Sequence-to-Sequence (Seq2Seq) models emerged as a substantial advancement in MT. These models utilize recurrent neural networks (RNNs) to convert sequences from the source to the target language \cite{bakarola2021attention}\cite{phua2022sequence}\cite{do2018sequence}\cite{sutskever2014sequence}. A significant enhancement to this approach was the introduction of attention mechanisms, allowing models to dynamically focus on different parts of the source sentence during decoding, significantly improving translation accuracy for longer sentences \cite{bahdanau2015neural}. The attention mechanism facilitated dynamic context generation, weighted sum calculations, and sophisticated decoding strategies, all contributing to heightened accuracy and interpretability in MT.

Hybrid models, which synergistically combine SMT and neural approaches, emerged as a potent solution. These models seek to harness the statistical robustness of SMT and the contextual nuances captured by neural networks, creating translations that are both semantically rich and syntactically correct.

The introduction of the Transformer architecture marked another paradigm shift in MT. It moved away from recurrent layers and emphasized self-attention mechanisms, allowing each word in a sequence to attend to all others \cite{han2021transformer}\cite{huang2022flowformer}\cite{vaswani2017attention}. Innovations like multi-head attention, feedforward neural networks, positional encoding, layer normalization, and residual connections enhanced the architecture's capabilities. Transformer-based models, such as BERT, excelled in various NLP tasks, including MT, often setting new benchmarks \cite{devlin2019bert}. Their parallel processing capabilities led to scalable models that established new standards in terms of BLEU scores and real-time translation.

In response to the need for continuous improvement, ensemble methods were revisited and applied to Transformer models. These ensemble approaches, which amalgamate multiple Transformer models or integrate Transformers with other architectures, consistently demonstrated improvements in translation quality and robustness over single-model approaches.

Despite the numerous advantages of Transformers, challenges such as memory-intensive computations and potential computational overhead for shorter sequences persist. However, the evolution of MT—from statistical methods to neural, ensemble, and hybrid techniques, culminating in Transformers—showcases the relentless pursuit of excellence in this field. The ongoing integration of ensemble and hybrid models, coupled with continuous advancements in core technologies, indicates a promising trajectory for enhancing the efficiency and adaptability of MT systems.

\subsection{Question Answering Systems}
Question Answering (QA) systems, meticulously designed to extract accurate answers from a plethora of structured or unstructured data sources in response to specific user queries, have undergone a substantial evolution over time \cite{hirschman2001natural}\cite{rajpurkar2016squad}\cite{allam2012question}\cite{ojokoh2018review}. This impressive progress can be chiefly attributed to groundbreaking innovations such as Information Retrieval (IR)-based QA systems and Attention-based Sequence-to-Sequence (Seq2Seq) models, each contributing unique strengths to the realm of QA.

IR-based QA systems are renowned for their proficiency in swiftly identifying, locating, and ranking pertinent documents or information segments from large-scale datasets without delving deeply into the semantics of the content \cite{manning2008introduction}\cite{bao2018information}\cite{sagrado2022information}. The mechanism begins with an efficient document indexing process, usually utilizing inverted indexing techniques to ensure rapid and accurate retrieval. Following this, the user's query undergoes analysis and transformation using advanced natural language processing (NLP) techniques to derive context and meaning. Relevant documents are retrieved based on the processed query, and algorithms such as term frequency-inverse document frequency (TF-IDF) and cosine similarity are applied to rank these documents based on relevance. The system then extracts potential answers from the top-ranked documents. IR-based systems stand out for their scalability and minimal training requirements, allowing them to be versatile across various domains. Additionally, the transparency in sourcing answers adds a layer of interpretability. However, these systems may face challenges in comprehending content semantics, providing precise granularity in answers, and may be influenced by the quality of the underlying data.

On the other hand, Attention-based Seq2Seq models, initially conceptualized for machine translation, have been extensively repurposed and optimized for QA tasks \cite{bahdanau2014neural}\cite{nie2017attention}\cite{afrae2020question}\cite{liu2019visual}. These models typically include an encoder, often implemented using Long Short-Term Memory (LSTM) or Gated Recurrent Unit (GRU), which processes the input sequence to generate context vectors. An attention mechanism is then deployed to assign weights to these vectors, thereby signifying the significance of each input token in determining the corresponding output token. The decoder, in turn, uses these weighted context vectors to generate the final output sequence. Attention-based Seq2Seq models have been pivotal in addressing several challenges in QA, such as processing extensive passages and enhancing answer precision, by dynamically focusing on relevant sections of the input text. These models have delivered state-of-the-art results across numerous QA benchmarks and showcased versatility across various domains and question types. Their architecture, celebrated for its capability to manage long sequences and its interpretability due to attention weights, is flexible enough to be fine-tuned or amalgamated into hybrid systems, thereby boosting their effectiveness. However, these models may pose computational challenges and are sometimes prone to overfitting on smaller datasets.

The incorporation of both IR-based and Attention-based Seq2Seq models into ensemble or hybrid systems signifies a strategic combination of diverse models aimed at capitalizing on their individual strengths while concurrently mitigating their respective weaknesses. Ensemble techniques such as bagging, boosting, or stacking can be employed to harmonize and aggregate outputs from different models, thereby bolstering predictive performance. For instance, a sophisticated hybrid ensemble model could synergistically combine the rapid retrieval capabilities of an IR-based system with the deep contextual understanding of an attention-based Seq2Seq model. This amalgamation would yield responses that are both precise and contextually rich. Such a system may also benefit from techniques like knowledge distillation, wherein the insights from a complex ensemble model are transferred to a smaller, more efficient model. This approach ensures an optimal balance between response speed and accuracy.

By actively embracing and integrating ensemble and hybrid methodologies, QA systems can continue to evolve, progressively refining their capacity to provide accurate and efficient responses to user queries. These combinations can lead to the development of QA systems that are not only robust and comprehensive but also capable of self-improvement and adaptation to the ever-evolving landscape of user needs and data complexities.

\subsection{Named Entity Recognition (NER)}
Named Entity Recognition (NER) stands as a pivotal task within the realm of Natural Language Processing (NLP), dedicated to categorizing specific instances of words or phrases, such as names of individuals, organizations, and geographical locations, within textual data into predefined classes\cite{sun2018overview}\cite{li2020survey}\cite{mohit2014named}\cite{nadeau2007survey}\cite{grishman1996message}. This task is central to numerous applications, including information retrieval, question answering, and relationship extraction, underscoring its importance in extracting structured information from unstructured text.

Historically, traditional models such as Conditional Random Fields (CRF) have been extensively utilized for NER tasks\cite{patil2020named}\cite{song2019named}\cite{khan2022named}\cite{yang2018conditional}\cite{lafferty2001conditional}. CRFs, being a type of discriminative probabilistic model, are effective for handling sequential data. They operate by modeling the conditional probability of output sequences (labels) given input sequences (words), thereby allowing context-sensitive predictions. The feature engineering process in CRFs is often intricate and involves crafting word-level features, linguistic features, and utilizing gazetteer lists and regular expressions to capture entity formats and other nuances in the data.

With the advent of deep learning, NER experienced a transformative shift towards more sophisticated models, which significantly reduced the reliance on manual feature engineering. Notable among these is the integration of Bi-directional Long Short-Term Memory networks (BiLSTMs) with a CRF layer, forming the BiLSTM-CRF model\cite{huang2015bidirectional}\cite{dai2019named}\cite{yan2021named}\cite{chenhao2019named}\cite{ma2016end}. This combination leverages the sequential memory and learning capabilities of BiLSTMs along with the sequence labeling strengths of CRFs. Such an architecture can effectively capture context from both directions (forward and backward) and model complex dependencies in data.

Moreover, the emergence of Transformer-based models, such as BERT, has further revolutionized the landscape of NER\cite{vaswani2017attention}\cite{evangelatos2021named}\cite{rouhou2022transformer}\cite{berragan2023transformer}\cite{devlin2019bert}. BERT, with its self-attention mechanisms and extensive pre-training on large corpora, facilitates fine-tuning for specific NER tasks. This approach has consistently demonstrated superior performance, as it encapsulates contextual information and intricacies of natural language, providing rich, dynamic representations.

In light of these advancements, there is a discernible trend towards the development and implementation of ensemble and hybrid models for NER. These methodologies aim to amalgamate various models, harnessing their respective strengths and compensating for their weaknesses. For instance, an ensemble model could combine the swift retrieval capabilities of a traditional CRF model, the contextual awareness of BiLSTM-CRF, and the extensive pre-trained knowledge of BERT. By doing so, ensemble and hybrid models strive to deliver enhanced predictive accuracy, robustness, and generalization across diverse domains and datasets.

In conclusion, while traditional models like CRFs and Support Vector Machines (SVMs) have exhibited commendable proficiency in NER tasks, the advent of more recent architectures such as BiLSTM-CRF and Transformer-based models has ushered in an era of elevated performance and minimized manual intervention. Ensemble and hybrid methodologies, by virtue of their ability to strategically combine and leverage different models, are emerging as promising frontiers, poised to further advance the field of NER.

\section{Challenges in Implementing Ensemble Deep Learning for NLP}
Ensemble deep learning, a sophisticated strategy that amalgamates the predictive power of multiple models, has carved out a significant niche in the realm of natural language processing (NLP) \cite{sutskever2014sequence}\cite{khaiphan2019deep}. This technique is engineered to bolster overall predictive performance by synthesizing insights and capabilities across diverse models. However, the deployment of ensemble deep learning brings to light a multitude of intricate challenges and nuanced considerations that warrant careful examination.

One of the foremost challenges is the substantial computational requirements associated with implementing ensemble deep learning methodologies. The intricate nature of combining multiple models necessitates robust computational infrastructures, often entailing high-performance GPUs or TPUs for efficient training. The simultaneous training and maintenance of diverse models can escalate the computational costs and extend the training duration, posing potential impediments for applications with real-time or time-sensitive demands \cite{sutskever2014sequence}\cite{khaiphan2019deep}. Furthermore, the risk of overfitting is amplified, particularly when the constituent models are closely correlated, leading to potentially inflated performance metrics that do not generalize well to unseen data \cite{opitz1999popular}. Ensuring heterogeneity amongst ensemble members is crucial to obviate this risk. This can be achieved through strategic deployment of disparate architectures, diverse initialization strategies, or variance in training data \cite{mueller2022artificial}.

The challenge of model interpretability surfaces prominently when deploying ensemble models. A confluence of multiple models tends to obscure the decision-making processes, complicating efforts to glean transparent insights \cite{caruana2015intelligible}. This opacity can be particularly problematic in domains where interpretability is paramount, such as in legal or healthcare settings. The intricacy of interpretation augments commensurately with an increase in ensemble size. Ascertaining an optimal ensemble size and adeptly selecting constituent models demands a judicious blend of expertise and experimental rigor \cite{hansen1990neural}. A diminutive ensemble may fall short of realizing tangible benefits, while an excessively large ensemble can precipitate computational conundrums.

Effective management of training data and ensuring diversity therein is a pivotal concern. Navigating through myriad linguistic variations and contexts mandates strategic planning to preclude biases and ensure representativeness \cite{chawla2002smote}. Addressing potential imbalances in data distribution across constituent models is imperative for preserving the equilibrium and efficacy of the ensemble. Moreover, the deployment and integration of ensemble models into production environments herald additional complexities. The endeavor necessitates meticulous engineering to guarantee efficient and harmonious interactions with other system components, alongside ensuring scalability to accommodate fluctuating workloads \cite{gupta2011scaling}.

Maintenance of ensemble models emerges as a continuous imperative. Given the mutable nature of data distributions, periodic reassessment and updating of individual models within an ensemble become crucial. Managing the requisite hardware and software resources for perpetually training and deploying these ensembles can indeed be resource-intensive and necessitate strategic planning. Furthermore, a persistent tension exists between the quest for interpretability and the pursuit of peak performance. Ensemble deep learning, while prioritizing performance, may inadvertently compromise on interpretability, ushering in ethical and practical quandaries in certain applications \cite{rudin2019stop}. Especially, for a ensemble of LLMs, the maintanance costs and budgeting on each individual candidate model, ranking module and fusion module place a great challenge as well. In light of these challenges, the deployment of ensemble deep learning in NLP warrants a holistic approach that judiciously balances computational demands, interpretability, and performance.

\section{Conclusion}
In the domain of Natural Language Processing (NLP), ensemble deep learning models have rapidly ascended as a formidable mechanism, recalibrating the thresholds of state-of-the-art performance by intricately navigating the multifaceted challenges inherent to human language. These ensemble models judiciously amalgamate the predictive prowess of multiple diverse models, thereby yielding a synergy that is capable of deciphering the myriad subtleties and complexities intrinsic to linguistic communication. The interplay between ensemble methods and deep learning architectures has been instrumental in sculpting the trajectory of advancements in NLP, fostering significant breakthroughs across an array of linguistic tasks.

The utilization of ensemble methods in NLP is characterized by a pragmatic confluence of individual model strengths, effectively counteracting their intrinsic limitations and culminating in an enhanced performance across a spectrum of complex linguistic tasks. From unraveling the complexities of machine translation and deciphering nuanced sentiments in textual data, to achieving unprecedented precision in tasks such as named entity recognition, ensemble techniques have consistently validated their indispensability. These methods, encompassing a variety of techniques such as voting, bagging, boosting, and stacking, have been meticulously adapted and tailored to meet the unique exigencies posed by the multifarious nuances of human language \cite{zhou2012ensemble}.

In the context of NLP, ensemble models transcend the conventional boundaries of accuracy enhancement, introducing an unparalleled degree of robustness and versatility in tackling linguistic challenges. Moreover, hybrid models, which seamlessly blend different learning paradigms, further augment the capabilities of ensemble techniques. By fusing traditional machine learning algorithms with sophisticated deep learning models, hybrid ensembles emerge as a holistic solution capable of harnessing complementary strengths and achieving robust performance across diverse NLP tasks.

For LLM ensemble, routing and blending methods represent powerful ensemble strategies that leverage the strengths of multiple LLMs to enhance performance across various tasks. By dynamically selecting or combining outputs from different models, these approaches ensure robustness and versatility. Additionally, the ensemble concept extends to optimizing serving costs, as seen in techniques like speculative decoding and Cascade Speculative Drafting, which focus on improving inference efficiency without compromising quality. However, these methods also present challenges, such as balancing trade-offs between performance and costs, as well as the interpretability.

In summation, the synergistic alliance between ensemble methods and deep learning models in the realm of NLP epitomizes the scientific community's unwavering endeavor to continually redefine the boundaries of linguistic understanding and computational capabilities. As the technological landscape evolves, marked by burgeoning computational prowess and incessant refinement of methodologies, it is envisaged that this symbiotic confluence will continue to catalyze groundbreaking advancements in deciphering and processing human language, thereby ushering in an era of unparalleled linguistic comprehension and interaction \cite{goodfellow2016deep}.

\bibliographystyle{unsrt}
\bibliography{reference}

\begin{thebibliography}{100}

\bibitem{jones1994natural}
K.~S. Jones.
\newblock {\em Natural language processing: a historical review}, pages 3--16.
\newblock 1994.

\bibitem{liddy2001natural}
E.~D. Liddy.
\newblock Natural language processing.
\newblock 2001.

\bibitem{jurafsky2014speech}
D.~Jurafsky and J.~H. Martin.
\newblock {\em Speech and Language Processing}.
\newblock 2014.

\bibitem{hutchins1995machine}
W.~J. Hutchins.
\newblock {\em Machine translation: A brief history}, pages 431--445.
\newblock Pergamon, 1995.

\bibitem{nadkarni2011natural}
P.~M. Nadkarni, L.~Ohno-Machado, and W.~W. Chapman.
\newblock Natural language processing: an introduction.
\newblock {\em Journal of the American Medical Informatics Association}, 18(5):544--551, 2011.

\bibitem{turing1950mind}
Alan~Mathison Turing.
\newblock Mind.
\newblock {\em Mind}, 59(236):433--460, 1950.

\bibitem{manning2008introduction}
C.~D. Manning, P.~Raghavan, and H.~Sch{\"u}tze.
\newblock {\em Introduction to Information Retrieval}.
\newblock Cambridge University Press, 2008.

\bibitem{deng2014deep}
L.~Deng and D.~Yu.
\newblock Deep learning: methods and applications.
\newblock {\em Foundations and trends{\textregistered} in signal processing}, 7(3–4):197--387, 2014.

\bibitem{zhang2021understanding}
C.~Zhang, S.~Bengio, M.~Hardt, B.~Recht, and O.~Vinyals.
\newblock Understanding deep learning (still) requires rethinking generalization.
\newblock {\em Communications of the ACM}, 64(3):107--115, 2021.

\bibitem{goodfellow2016deep}
Ian Goodfellow, Yoshua Bengio, and Aaron Courville.
\newblock {\em Deep learning}.
\newblock MIT press, 2016.

\bibitem{vaswani2017attention}
A.~Vaswani, N.~Shazeer, N.~Parmar, J.~Uszkoreit, L.~Jones, A.~N. Gomez, et~al.
\newblock Attention is all you need.
\newblock pages 5998--6008, 2017.

\bibitem{mikolov2013efficient}
T.~Mikolov, K.~Chen, G.~Corrado, and J.~Dean.
\newblock Efficient estimation of word representations in vector space.
\newblock {\em arXiv preprint arXiv:1301.3781}, 2013.

\bibitem{howard2018universal}
Jeremy Howard and Sebastian Ruder.
\newblock Universal language model fine-tuning for text classification.
\newblock {\em arXiv preprint arXiv:1801.06146}, 2018.

\bibitem{young2018recent}
Tom Young, Devamanyu Hazarika, Soujanya Poria, and Erik Cambria.
\newblock Recent trends in deep learning based natural language processing.
\newblock {\em ieee Computational intelligenCe magazine}, 13(3):55--75, 2018.

\bibitem{devlin2019bert}
J.~Devlin, M.~W. Chang, K.~Lee, and K.~Toutanova.
\newblock Bert: Pre-training of deep bidirectional transformers for language understanding.
\newblock In {\em Proceedings of the 2019 Conference of the North American Chapter of the Association for Computational Linguistics: Human Language Technologies}, pages 4171--4186, 2019.

\bibitem{brown2020languagemodelsfewshotlearners}
Tom~B. Brown, Benjamin Mann, Nick Ryder, Melanie Subbiah, Jared Kaplan, Prafulla Dhariwal, Arvind Neelakantan, Pranav Shyam, Girish Sastry, Amanda Askell, Sandhini Agarwal, Ariel Herbert-Voss, Gretchen Krueger, Tom Henighan, Rewon Child, Aditya Ramesh, Daniel~M. Ziegler, Jeffrey Wu, Clemens Winter, Christopher Hesse, Mark Chen, Eric Sigler, Mateusz Litwin, Scott Gray, Benjamin Chess, Jack Clark, Christopher Berner, Sam McCandlish, Alec Radford, Ilya Sutskever, and Dario Amodei.
\newblock Language models are few-shot learners, 2020.

\bibitem{colombo2024saullm7bpioneeringlargelanguage}
Pierre Colombo, Telmo~Pessoa Pires, Malik Boudiaf, Dominic Culver, Rui Melo, Caio Corro, Andre F.~T. Martins, Fabrizio Esposito, Vera~Lúcia Raposo, Sofia Morgado, and Michael Desa.
\newblock Saullm-7b: A pioneering large language model for law, 2024.

\bibitem{cui2024chatlawmultiagentcollaborativelegal}
Jiaxi Cui, Munan Ning, Zongjian Li, Bohua Chen, Yang Yan, Hao Li, Bin Ling, Yonghong Tian, and Li~Yuan.
\newblock Chatlaw: A multi-agent collaborative legal assistant with knowledge graph enhanced mixture-of-experts large language model, 2024.

\bibitem{bender2021dangers}
Emily~M. Bender, Timnit Gebru, Angelina McMillan-Major, and Shmargaret Shmitchell.
\newblock On the dangers of stochastic parrots: Can language models be too big?
\newblock In {\em Proceedings of the 2021 {ACM} Conference on Fairness, Accountability, and Transparency}, New York, March 2021. Association for Computer Machinery – {ACM}.

\bibitem{dietterich2000ensemble}
T.~G. Dietterich.
\newblock Ensemble methods in machine learning.
\newblock In {\em International workshop on multiple classifier systems}, pages 1--15. Springer Berlin Heidelberg, 2000.

\bibitem{opitz1999popular}
D.~Opitz and R.~Maclin.
\newblock Popular ensemble methods: An empirical study.
\newblock {\em Journal of Artificial Intelligence Research}, 11:169--198, 1999.

\bibitem{zhou2012ensemble}
Zhi-Hua Zhou.
\newblock {\em Ensemble methods: foundations and algorithms}.
\newblock CRC press, 2012.

\bibitem{mi2019bagging}
X.~Mi, F.~Zou, and R.~Zhu.
\newblock Bagging and deep learning in optimal individualized treatment rules.
\newblock {\em Biometrics}, 75(2):674--684, 2019.

\bibitem{kumar2008from}
L.~Kumar and D.~P. Sinha.
\newblock From cmp to crs: an overview of stacking techniques of seismic data.
\newblock In {\em 7th Biennial international conference and exposition on petroleum geophysics}, volume 414, 2008.

\bibitem{breiman1996bagging}
Leo Breiman.
\newblock Bagging predictors.
\newblock {\em Machine learning}, 24:123--140, 1996.

\bibitem{freund1997decision}
Yoav Freund and Robert~E Schapire.
\newblock A decision-theoretic generalization of on-line learning and an application to boosting.
\newblock {\em Journal of computer and system sciences}, 55(1):119--139, 1997.

\bibitem{wolpert1992stacked}
David~H Wolpert.
\newblock Stacked generalization.
\newblock {\em Neural networks}, 5(2):241--259, 1992.

\bibitem{kurniasari2020sentiment}
L.~Kurniasari and A.~Setyanto.
\newblock Sentiment analysis using recurrent neural network.
\newblock In {\em Journal of Physics: Conference Series}, volume 1471, page 012018. IOP Publishing, 2020.

\bibitem{arras2017explaining}
L.~Arras, G.~Montavon, K.~R. M{\"u}ller, and W.~Samek.
\newblock Explaining recurrent neural network predictions in sentiment analysis.
\newblock {\em arXiv preprint arXiv:1706.07206}, 2017.

\bibitem{patel2019sentiment}
A.~Patel and A.~K. Tiwari.
\newblock Sentiment analysis by using recurrent neural network.
\newblock In {\em Proceedings of 2nd International Conference on Advanced Computing and Software Engineering (ICACSE)}, February 2019.

\bibitem{tembhurne2021sentiment}
J.~V. Tembhurne and T.~Diwan.
\newblock Sentiment analysis in textual, visual and multimodal inputs using recurrent neural networks.
\newblock {\em Multimedia Tools and Applications}, 80:6871--6910, 2021.

\bibitem{alhumoud2022arabic}
S.~O. Alhumoud and A.~A. Al~Wazrah.
\newblock Arabic sentiment analysis using recurrent neural networks: a review.
\newblock {\em Artificial Intelligence Review}, 55(1):707--748, 2022.

\bibitem{mao2022review}
S.~Mao and E.~Sejdić.
\newblock A review of recurrent neural network-based methods in computational physiology.
\newblock {\em IEEE Transactions on Neural Networks and Learning Systems}, 2022.

\bibitem{elman1990finding}
J.~L. Elman.
\newblock Finding structure in time.
\newblock {\em Cognitive science}, 14(2):179--211, 1990.

\bibitem{graves2012supervised}
A.~Graves.
\newblock {\em Supervised sequence labelling with recurrent neural networks}.
\newblock Springer Science \& Business Media, 2012.

\bibitem{lipton2015critical}
Z.~C. Lipton, J.~Berkowitz, and C.~Elkan.
\newblock A critical review of recurrent neural networks for sequence learning.
\newblock {\em arXiv preprint arXiv:1506.00019}, 2015.

\bibitem{ming2017understanding}
Y.~Ming, S.~Cao, R.~Zhang, Z.~Li, Y.~Chen, Y.~Song, and H.~Qu.
\newblock Understanding hidden memories of recurrent neural networks.
\newblock In {\em 2017 IEEE conference on visual analytics science and technology (VAST)}, pages 13--24. IEEE, 2017.

\bibitem{werbos1990backpropagation}
P.~J. Werbos.
\newblock Backpropagation through time: what it does and how to do it.
\newblock {\em Proceedings of the IEEE}, 78(10):1550--1560, 1990.

\bibitem{zhang2018character}
X.~Zhang, J.~Zhao, and Y.~LeCun.
\newblock Character-level convolutional networks for text classification.
\newblock In {\em Proceedings of the 28th International Conference on Neural Information Processing Systems - Volume 1}, 2018.

\bibitem{baktha2017investigation}
K.~Baktha and B.~K. Tripathy.
\newblock Investigation of recurrent neural networks in the field of sentiment analysis.
\newblock In {\em 2017 International Conference on Communication and Signal Processing (ICCSP)}, pages 2047--2050. IEEE, 2017.

\bibitem{sachin2020sentiment}
S.~Sachin, A.~Tripathi, N.~Mahajan, S.~Aggarwal, and P.~Nagrath.
\newblock Sentiment analysis using gated recurrent neural networks.
\newblock {\em SN Computer Science}, 1:1--13, 2020.

\bibitem{islam2020review}
S.~Islam, N.~Ab~Ghani, and M.~Ahmed.
\newblock A review on recent advances in deep learning for sentiment analysis: Performances, challenges and limitations.
\newblock {\em Compusoft}, 9(7):3775--3783, 2020.

\bibitem{maheswaranathan2020how}
N.~Maheswaranathan and D.~Sussillo.
\newblock How recurrent networks implement contextual processing in sentiment analysis.
\newblock {\em arXiv preprint arXiv:2004.08013}, 2020.

\bibitem{ito2020contextual}
T.~Ito, K.~Tsubouchi, H.~Sakaji, T.~Yamashita, and K.~Izumi.
\newblock Contextual sentiment neural network for document sentiment analysis.
\newblock {\em Data Science and Engineering}, 5:180--192, 2020.

\bibitem{baziotis2017datastories}
C.~Baziotis, N.~Pelekis, and C.~Doulkeridis.
\newblock Datastories at semeval-2017 task 4: Deep lstm with attention for message-level and topic-based sentiment analysis.
\newblock In {\em Proceedings of the 11th international workshop on semantic evaluation (SemEval-2017)}, pages 747--754, 2017.

\bibitem{birjali2021comprehensive}
M.~Birjali, M.~Kasri, and A.~Beni-Hssane.
\newblock A comprehensive survey on sentiment analysis: Approaches, challenges and trends.
\newblock {\em Knowledge-Based Systems}, 226:107134, 2021.

\bibitem{gandhi2021sentiment}
U.~D. Gandhi, P.~Malarvizhi~Kumar, G.~Chandra~Babu, and G.~Karthick.
\newblock Sentiment analysis on twitter data by using convolutional neural network (cnn) and long short term memory (lstm).
\newblock {\em Wireless Personal Communications}, pages 1--10, 2021.

\bibitem{rani2019deep}
S.~Rani and P.~Kumar.
\newblock Deep learning based sentiment analysis using convolution neural network.
\newblock {\em Arabian Journal for Science and Engineering}, 44:3305--3314, 2019.

\bibitem{yu2016visual}
Y.~Yu, H.~Lin, J.~Meng, and Z.~Zhao.
\newblock Visual and textual sentiment analysis of a microblog using deep convolutional neural networks.
\newblock {\em Algorithms}, 9(2):41, 2016.

\bibitem{kim2019sentiment}
H.~Kim and Y.~S. Jeong.
\newblock Sentiment classification using convolutional neural networks.
\newblock {\em Applied Sciences}, 9(11):2347, 2019.

\bibitem{kim2014convolutional}
Y.~Kim.
\newblock Convolutional neural networks for sentence classification.
\newblock {\em arXiv preprint arXiv:1408.5882}, 2014.

\bibitem{lecun2015deep}
Y.~LeCun, Y.~Bengio, and G.~Hinton.
\newblock Deep learning.
\newblock {\em Nature}, 521(7553):436--444, 2015.

\bibitem{zhang2015character}
X.~Zhang, J.~Zhao, and Y.~LeCun.
\newblock Character-level convolutional networks for text classification.
\newblock 28, 2015.

\bibitem{miedema2018sentiment}
F.~Miedema and S.~Bhulai.
\newblock Sentiment analysis with long short-term memory networks.
\newblock {\em Vrije Universiteit Amsterdam}, 1:1--17, 2018.

\bibitem{shobana2021efficient}
J.~Shobana and M.~Murali.
\newblock An efficient sentiment analysis methodology based on long short-term memory networks.
\newblock {\em Complex \& Intelligent Systems}, 7(5):2485--2501, 2021.

\bibitem{al2019using}
Mohammad Al-Smadi, Bashar Talafha, Mahmoud Al-Ayyoub, and Yaser Jararweh.
\newblock Using long short-term memory deep neural networks for aspect-based sentiment analysis of arabic reviews.
\newblock {\em International Journal of Machine Learning and Cybernetics}, 10:2163--2175, 2019.

\bibitem{muhammad2021sentiment}
P.~F. Muhammad, R.~Kusumaningrum, and A.~Wibowo.
\newblock Sentiment analysis using word2vec and long short-term memory (lstm) for indonesian hotel reviews.
\newblock In {\em Procedia Computer Science}, volume 179, pages 728--735, 2021.

\bibitem{yadav2023habitat}
Karmesh Yadav, Ram Ramrakhya, Santhosh~Kumar Ramakrishnan, Theo Gervet, John Turner, Aaron Gokaslan, Noah Maestre, Angel~Xuan Chang, Dhruv Batra, Manolis Savva, et~al.
\newblock Habitat-matterport 3d semantics dataset.
\newblock In {\em Proceedings of the IEEE/CVF Conference on Computer Vision and Pattern Recognition}, pages 4927--4936, 2023.

\bibitem{staudemeyer2019understanding}
R.~C. Staudemeyer and E.~R. Morris.
\newblock Understanding lstm--a tutorial into long short-term memory recurrent neural networks.
\newblock {\em arXiv preprint arXiv:1909.09586}, 2019.

\bibitem{olah2015understanding}
C.~Olah.
\newblock Understanding lstm networks, 2015.

\bibitem{fei2018bidirectional}
H.~Fei and F.~Tan.
\newblock Bidirectional grid long short-term memory (bigridlstm): A method to address context-sensitivity and vanishing gradient.
\newblock {\em Algorithms}, 11(11):172, 2018.

\bibitem{hassan2017deep}
A.~Hassan and A.~Mahmood.
\newblock Deep learning approach for sentiment analysis of short texts.
\newblock In {\em 2017 3rd international conference on control, automation and robotics (ICCAR)}, pages 705--710. IEEE, 2017.

\bibitem{hochreiter1997long}
S.~Hochreiter and J.~Schmidhuber.
\newblock Long short-term memory.
\newblock {\em Neural computation}, 9(8):1735--1780, 1997.

\bibitem{Yin2017}
Mathieu Cliche.
\newblock Bb\_twtr at semeval-2017 task 4: Twitter sentiment analysis with cnns and lstms.
\newblock In {\em Proceedings of the 11th International Workshop on Semantic Evaluations (SemEval-2017)}, pages 573--580. Association for Computational Linguistics, 2017.

\bibitem{alaparthi2020bert}
S.~Alaparthi and M.~Mishra.
\newblock Bidirectional encoder representations from transformers (bert): A sentiment analysis odyssey.
\newblock {\em arXiv preprint arXiv:2007.01127}, 2020.

\bibitem{deepa2021bidirectional}
M.~D. Deepa.
\newblock Bidirectional encoder representations from transformers (bert) language model for sentiment analysis task.
\newblock {\em Turkish Journal of Computer and Mathematics Education (TURCOMAT)}, 12(7):1708--1721, 2021.

\bibitem{devlin2018bert}
Jacob Devlin, Ming-Wei Chang, Kenton Lee, and Kristina Toutanova.
\newblock Bert: Pre-training of deep bidirectional transformers for language understanding.
\newblock {\em arXiv preprint arXiv:1810.04805}, 2018.

\bibitem{sun2019utilizing}
Chi Sun, Luyao Huang, and Xipeng Qiu.
\newblock Utilizing bert for aspect-based sentiment analysis via constructing auxiliary sentence.
\newblock {\em arXiv preprint arXiv:1903.09588}, 2019.

\bibitem{kashyap2023gptneocommonsensereasoning}
Rohan Kashyap, Vivek Kashyap, and Narendra~C. P.
\newblock Gpt-neo for commonsense reasoning -- a theoretical and practical lens, 2023.

\bibitem{llama}
Hugo Touvron, Thibaut Lavril, Gautier Izacard, Xavier Martinet, Marie-Anne Lachaux, Timothée Lacroix, Baptiste Rozière, Naman Goyal, Eric Hambro, Faisal Azhar, Aurelien Rodriguez, Armand Joulin, Edouard Grave, and Guillaume Lample.
\newblock Llama: Open and efficient foundation language models, 2023.

\bibitem{mistral}
Albert~Q. Jiang, Alexandre Sablayrolles, Antoine Roux, Arthur Mensch, Blanche Savary, Chris Bamford, Devendra~Singh Chaplot, Diego de~las Casas, Emma~Bou Hanna, Florian Bressand, Gianna Lengyel, Guillaume Bour, Guillaume Lample, Lélio~Renard Lavaud, Lucile Saulnier, Marie-Anne Lachaux, Pierre Stock, Sandeep Subramanian, Sophia Yang, Szymon Antoniak, Teven~Le Scao, Théophile Gervet, Thibaut Lavril, Thomas Wang, Timothée Lacroix, and William~El Sayed.
\newblock Mixtral of experts, 2024.

\bibitem{jiang2023llmblenderensemblinglargelanguage}
Dongfu Jiang, Xiang Ren, and Bill~Yuchen Lin.
\newblock Llm-blender: Ensembling large language models with pairwise ranking and generative fusion, 2023.

\bibitem{chen2023detectingerrorsestimatingaccuracy}
Jiefeng Chen, Frederick Liu, Besim Avci, Xi~Wu, Yingyu Liang, and Somesh Jha.
\newblock Detecting errors and estimating accuracy on unlabeled data with self-training ensembles, 2023.

\bibitem{chen2023frugalgptuselargelanguage}
Lingjiao Chen, Matei Zaharia, and James Zou.
\newblock Frugalgpt: How to use large language models while reducing cost and improving performance, 2023.

\bibitem{shnitzer2023largelanguagemodelrouting}
Tal Shnitzer, Anthony Ou, Mírian Silva, Kate Soule, Yuekai Sun, Justin Solomon, Neil Thompson, and Mikhail Yurochkin.
\newblock Large language model routing with benchmark datasets, 2023.

\bibitem{leviathan2023fastinferencetransformersspeculative}
Yaniv Leviathan, Matan Kalman, and Yossi Matias.
\newblock Fast inference from transformers via speculative decoding, 2023.

\bibitem{chen2024cascadespeculativedraftingfaster}
Ziyi Chen, Xiaocong Yang, Jiacheng Lin, Chenkai Sun, Kevin Chen-Chuan Chang, and Jie Huang.
\newblock Cascade speculative drafting for even faster llm inference, 2024.

\bibitem{niessen2000evaluation}
S.~Nießen, F.~J. Och, G.~Leusch, and H.~Ney.
\newblock An evaluation tool for machine translation: Fast evaluation for mt research.
\newblock In {\em LREC}, May 2000.

\bibitem{kenny2018machine}
D.~Kenny.
\newblock {\em Machine translation}, pages 428--445.
\newblock Routledge, 2018.

\bibitem{sutskever2014sequence}
I.~Sutskever, O.~Vinyals, and Q.~V. Le.
\newblock Sequence to sequence learning with neural networks.
\newblock pages 3104--3112, 2014.

\bibitem{hearne2011statistical}
M.~Hearne and A.~Way.
\newblock Statistical machine translation: a guide for linguists and translators.
\newblock {\em Language and Linguistics Compass}, 5(5):205--226, 2011.

\bibitem{lopez2008statistical}
A.~Lopez.
\newblock Statistical machine translation.
\newblock {\em ACM Computing Surveys (CSUR)}, 40(3):1--49, 2008.

\bibitem{mahata2019mtil2017}
S.~K. Mahata, D.~Das, and S.~Bandyopadhyay.
\newblock Mtil2017: Machine translation using recurrent neural network on statistical machine translation.
\newblock {\em Journal of Intelligent Systems}, 28(3):447--453, 2019.

\bibitem{doherty2012taking}
S.~Doherty, D.~Kenny, and A.~Way.
\newblock Taking statistical machine translation to the student translator.
\newblock 2012.

\bibitem{brown1993mathematics}
P.~F. Brown, V.~J.~D. Pietra, S.~A.~D. Pietra, and R.~L. Mercer.
\newblock The mathematics of statistical machine translation: Parameter estimation.
\newblock {\em Computational Linguistics}, 19(2):263--311, 1993.

\bibitem{bakarola2021attention}
V.~Bakarola and J.~Nasriwala.
\newblock Attention based sequence to sequence learning for machine translation of low resourced indic languages--a case of sanskrit to hindi.
\newblock {\em arXiv preprint arXiv:2110.00435}, 2021.

\bibitem{phua2022sequence}
Y.~T. Phua, S.~Navaratnam, C.~M. Kang, and W.~S. Che.
\newblock Sequence-to-sequence neural machine translation for english-malay.
\newblock {\em IAES International Journal of Artificial Intelligence}, 11(2):658, 2022.

\bibitem{do2018sequence}
Q.~T. Do, S.~Sakti, and S.~Nakamura.
\newblock Sequence-to-sequence models for emphasis speech translation.
\newblock {\em IEEE/ACM Transactions on Audio, Speech, and Language Processing}, 26(10):1873--1883, 2018.

\bibitem{bahdanau2015neural}
D.~Bahdanau, K.~Cho, and Y.~Bengio.
\newblock Neural machine translation by jointly learning to align and translate.
\newblock In {\em Proceedings of the International Conference on Learning Representations (ICLR)}, 2015.

\bibitem{han2021transformer}
K.~Han, A.~Xiao, E.~Wu, J.~Guo, C.~Xu, and Y.~Wang.
\newblock Transformer in transformer.
\newblock {\em Advances in Neural Information Processing Systems}, 34:15908--15919, 2021.

\bibitem{huang2022flowformer}
Z.~Huang, X.~Shi, C.~Zhang, Q.~Wang, K.~C. Cheung, H.~Qin, et~al.
\newblock Flowformer: A transformer architecture for optical flow.
\newblock In {\em European Conference on Computer Vision}, pages 668--685. Springer Nature Switzerland, 2022.

\bibitem{hirschman2001natural}
Lynette Hirschman and Robert Gaizauskas.
\newblock Natural language question answering: the view from here.
\newblock {\em natural language engineering}, 7(4):275--300, 2001.

\bibitem{rajpurkar2016squad}
Pranav Rajpurkar, Jian Zhang, Konstantin Lopyrev, and Percy Liang.
\newblock Squad: 100,000+ questions for machine comprehension of text.
\newblock {\em arXiv preprint arXiv:1606.05250}, 2016.

\bibitem{allam2012question}
A.~M.~N. Allam and M.~H. Haggag.
\newblock The question answering systems: A survey.
\newblock {\em International Journal of Research and Reviews in Information Sciences (IJRRIS)}, 2(3), 2012.

\bibitem{ojokoh2018review}
B.~Ojokoh and E.~Adebisi.
\newblock A review of question answering systems.
\newblock {\em Journal of Web Engineering}, 17(8):717--758, 2018.

\bibitem{bao2018information}
J.~Bao, N.~Duan, M.~Zhou, and T.~Zhao.
\newblock An information retrieval-based approach to table-based question answering.
\newblock In {\em Natural Language Processing and Chinese Computing: 6th CCF International Conference, NLPCC 2017}, pages 601--611. Springer International Publishing, 2018.

\bibitem{sagrado2022information}
A.~Sagrado~Sala.
\newblock Master dissertation: Information retrieval for question answering based on distributed representations.
\newblock Master's thesis, 2022.

\bibitem{bahdanau2014neural}
D.~Bahdanau, K.~Cho, and Y.~Bengio.
\newblock Neural machine translation by jointly learning to align and translate.
\newblock {\em arXiv preprint arXiv:1409.0473}, 2014.

\bibitem{nie2017attention}
Y.~P. Nie, Y.~Han, J.~M. Huang, B.~Jiao, and A.~P. Li.
\newblock Attention-based encoder-decoder model for answer selection in question answering.
\newblock {\em Frontiers of Information Technology \& Electronic Engineering}, 18(4):535--544, 2017.

\bibitem{afrae2020question}
B.~Afrae, B.~A. Mohamed, and A.~A. Boudhir.
\newblock A question answering system with a sequence to sequence grammatical correction.
\newblock In {\em Proceedings of the 3rd International Conference on Networking, Information Systems \& Security}, pages 1--6, 2020.

\bibitem{liu2019visual}
Y.~Liu, X.~Zhang, F.~Huang, X.~Tang, and Z.~Li.
\newblock Visual question answering via attention-based syntactic structure tree-lstm.
\newblock {\em Applied Soft Computing}, 82:105584, 2019.

\bibitem{sun2018overview}
P.~Sun, X.~Yang, X.~Zhao, and Z.~Wang.
\newblock An overview of named entity recognition.
\newblock In {\em 2018 International Conference on Asian Language Processing (IALP)}, pages 273--278. IEEE, 2018.

\bibitem{li2020survey}
J.~Li, A.~Sun, J.~Han, and C.~Li.
\newblock A survey on deep learning for named entity recognition.
\newblock {\em IEEE Transactions on Knowledge and Data Engineering}, 34(1):50--70, 2020.

\bibitem{mohit2014named}
B.~Mohit.
\newblock Named entity recognition.
\newblock In {\em Natural language processing of semitic languages}, pages 221--245. Springer Berlin Heidelberg, 2014.

\bibitem{nadeau2007survey}
David Nadeau and Satoshi Sekine.
\newblock A survey of named entity recognition and classification.
\newblock {\em Lingvisticae Investigationes}, 30(1):3--26, 2007.

\bibitem{grishman1996message}
Ralph Grishman and Beth~M Sundheim.
\newblock Message understanding conference-6: A brief history.
\newblock In {\em COLING 1996 Volume 1: The 16th International Conference on Computational Linguistics}, 1996.

\bibitem{patil2020named}
N.~Patil, A.~Patil, and B.~V. Pawar.
\newblock Named entity recognition using conditional random fields.
\newblock {\em Procedia Computer Science}, 167:1181--1188, 2020.

\bibitem{song2019named}
S.~Song, N.~Zhang, and H.~Huang.
\newblock Named entity recognition based on conditional random fields.
\newblock {\em Cluster Computing}, 22:5195--5206, 2019.

\bibitem{khan2022named}
W.~Khan, A.~Daud, K.~Shahzad, T.~Amjad, A.~Banjar, and H.~Fasihuddin.
\newblock Named entity recognition using conditional random fields.
\newblock {\em Applied Sciences}, 12(13):6391, 2022.

\bibitem{yang2018conditional}
Xiaoran Yang and Wenkang Huang.
\newblock A conditional random fields approach to clinical name entity recognition.
\newblock In {\em CCKS tasks}, pages 1--6, 2018.

\bibitem{lafferty2001conditional}
J.~Lafferty, A.~McCallum, and F.~C. Pereira.
\newblock Conditional random fields: Probabilistic models for segmenting and labeling sequence data.
\newblock In {\em Proceedings of the 18th International Conference on Machine Learning}, pages 282--289, 2001.

\bibitem{huang2015bidirectional}
Z.~Huang, W.~Xu, and K.~Yu.
\newblock Bidirectional lstm-crf models for sequence tagging.
\newblock {\em arXiv preprint arXiv:1508.01991}, 2015.

\bibitem{dai2019named}
Z.~Dai, X.~Wang, P.~Ni, Y.~Li, G.~Li, and X.~Bai.
\newblock Named entity recognition using bert bilstm crf for chinese electronic health records.
\newblock In {\em 2019 12th International Congress on Image and Signal Processing, Biomedical Engineering and Informatics (CISP-BMEI)}, pages 1--5. IEEE, 2019.

\bibitem{yan2021named}
Rongen Yan, Xue Jiang, and Depeng Dang.
\newblock Named entity recognition by using xlnet-bilstm-crf.
\newblock {\em Neural Processing Letters}, 53(5):3339--3356, 2021.

\bibitem{chenhao2019named}
Z.~Chenhao and W.~Chengyao.
\newblock Named entity recognition in steel field based on bilstm-crf model.
\newblock In {\em Journal of Physics: Conference Series}, volume 1314, page 012217. IOP Publishing, 2019.

\bibitem{ma2016end}
X.~Ma and E.~Hovy.
\newblock End-to-end sequence labeling via bi-directional lstm-cnns-crf.
\newblock In {\em Proceedings of the 54th Annual Meeting of the Association for Computational Linguistics (Volume 1: Long Papers)}, pages 1064--1074, 2016.

\bibitem{evangelatos2021named}
P.~Evangelatos, C.~Iliou, T.~Mavropoulos, K.~Apostolou, T.~Tsikrika, S.~Vrochidis, and I.~Kompatsiaris.
\newblock Named entity recognition in cyber threat intelligence using transformer-based models.
\newblock In {\em 2021 IEEE International Conference on Cyber Security and Resilience (CSR)}, pages 348--353. IEEE, 2021.

\bibitem{rouhou2022transformer}
A.~C. Rouhou, M.~Dhiaf, Y.~Kessentini, and S.~B. Salem.
\newblock Transformer-based approach for joint handwriting and named entity recognition in historical document.
\newblock {\em Pattern Recognition Letters}, 155:128--134, 2022.

\bibitem{berragan2023transformer}
C.~Berragan, A.~Singleton, A.~Calafiore, and J.~Morley.
\newblock Transformer based named entity recognition for place name extraction from unstructured text.
\newblock {\em International Journal of Geographical Information Science}, 37(4):747--766, 2023.

\bibitem{khaiphan2019deep}
T.~Khai~Tran and T.~Thi~Phan.
\newblock Deep learning application to ensemble learning—the simple, but effective, approach to sentiment classifying.
\newblock {\em Applied Sciences}, 9(13):2760, 2019.

\bibitem{mueller2022artificial}
B.~Mueller, T.~Kinoshita, A.~Peebles, M.~A. Graber, and S.~Lee.
\newblock Artificial intelligence and machine learning in emergency medicine: a narrative review.
\newblock {\em Acute medicine \& surgery}, 9(1):e740, 2022.

\bibitem{caruana2015intelligible}
R.~Caruana, Y.~Lou, J.~Gehrke, P.~Koch, M.~Sturm, and N.~Elhadad.
\newblock Intelligible models for healthcare: Predicting pneumonia risk and hospital 30-day readmission.
\newblock In {\em Proceedings of the 21th ACM SIGKDD International Conference on Knowledge Discovery and Data Mining}, 2015.

\bibitem{hansen1990neural}
L.~K. Hansen and P.~Salamon.
\newblock Neural network ensembles.
\newblock {\em IEEE transactions on pattern analysis and machine intelligence}, 12(10):993--1001, 1990.

\bibitem{chawla2002smote}
N.~V. Chawla, K.~W. Bowyer, L.~O. Hall, and W.~P. Kegelmeyer.
\newblock Smote: synthetic minority over-sampling technique.
\newblock {\em Journal of artificial intelligence research}, 16:321--357, 2002.

\bibitem{gupta2011scaling}
A.~Gupta and A.~Singhal.
\newblock Scaling and performance of the apache solr search engine.
\newblock In {\em Proceedings of the 16th international conference on World Wide Web}, 2011.

\bibitem{rudin2019stop}
C.~Rudin.
\newblock Stop explaining black box machine learning models for high stakes decisions and use interpretable models instead.
\newblock {\em Nature Machine Intelligence}, 1(5):206--215, 2019.

\end{thebibliography}

\appendix
\end{document}